\documentclass{article}

\usepackage[workshop, final]{neurips_2026}

\usepackage[utf8]{inputenc} 
\usepackage[T1]{fontenc}    
\usepackage{hyperref}       
\usepackage{url}            
\usepackage{booktabs}       
\usepackage{amsfonts}       
\usepackage{nicefrac}       
\usepackage{microtype}      
\usepackage{xcolor}         
\usepackage{graphicx}
\usepackage{multirow}
\usepackage{float}

\title{Behavioral Forgetting Is Not Privacy in Machine Unlearning}

\author{ Tyler Lizzo, Larry Heck\\ AI Virtual Assistant (AVA) Lab,
   Georgia Institute of Technology  \\ \texttt{\{lizzo,larryheck\}@gatech.edu}}

\begin{document}

\maketitle

\begin{abstract}
Machine unlearning aims to remove targeted information from large language models, but existing evaluations often measure forgetting only on a designated benchmark distribution. We argue that such behavioral forgetting is an insufficient standard for privacy-oriented unlearning, and that evaluation should instead test whether forgetting persists across the access paths through which the same information may be recovered. We demonstrate this with a controlled case study of cross-lingual querying: across multiple multilingual model families, forgetting varies substantially when semantically equivalent information is queried in different languages and scripts. A single access path is therefore enough to expose a real gap between behavioral forgetting and information inaccessibility, motivating broader evaluation of access-path generalization and adversarial recoverability alongside in-distribution forgetting.
\end{abstract}

\section{Introduction}

Large language models (LLMs) increasingly internalize personal information, copyrighted content, proprietary data, and other sensitive knowledge, motivating interest in \emph{machine unlearning}: the selective removal of targeted information from a deployed model. Unlike retraining from scratch, practical unlearning methods attempt to remove this information while preserving the model's broader capabilities. However, a central challenge remains: determining whether targeted information has actually been removed. Existing evaluations rely primarily on behavioral probes, benchmark performance, or statistical tests, which can demonstrate changes in model behavior but cannot establish that forgotten information is unrecoverable \citep{jagielski2023measuringforgettingmemorizedtraining}.

This distinction is particularly important for privacy. Successful forgetting under a designated evaluation distribution does not necessarily imply that the underlying information is inaccessible through other queries or representations. We therefore distinguish \emph{behavioral forgetting}, whether a model appears to forget under a specified evaluation distribution, from the stronger privacy objective of \emph{information inaccessibility}, which we treat as non-recoverability under a specified family of access mechanisms or threat model. Empirical evaluation cannot certify the latter, but it can provide stronger evidence by testing whether forgetting remains robust across alternative ways of accessing the same information.

Multilingual LLMs provide a natural setting for exposing this gap. Semantically equivalent information can be queried across different languages and scripts while the underlying information remains approximately fixed. If information appears forgotten in one language but remains recoverable in another, then a monolingual evaluation can overstate the strength of forgetting. We refer to these alternative mechanisms for querying or recovering the same information as \emph{access paths}. Cross-lingual querying is one such access path: it changes the surface form through which targeted information is queried while approximately preserving the underlying information itself, and multilingual factual associations provide a controlled setting in which this can be tested directly. A single access path is enough to test our position; we use cross-lingual unlearning as that case study and take up broader access paths in Section~\ref{sec:privacy_framework}.

Our position is that behavioral forgetting under a fixed evaluation distribution is insufficient evidence for privacy-oriented unlearning. We advance this position in three ways. First, we distinguish behavioral forgetting from the stronger objective of information inaccessibility and introduce access-path evaluation as a principle for testing whether forgetting persists beyond the original evaluation distribution. Second, we demonstrate this directly with cross-lingual querying: using results from cTOFU, we show that conclusions about unlearning success can change within this single, controlled access path. Third, we argue that this is sufficient evidence to motivate broader future work in which privacy-oriented benchmarks progress from in-distribution forgetting to access-path generalization and adversarial recoverability.

A reasonable alternative is that fixed-distribution evaluation is a practical and sufficient standard when broader testing is costly. We argue this supports only a narrower claim: success on a designated benchmark, not evidence that information is broadly inaccessible. A single access path where forgetting fails to generalize is already enough to show fixed-distribution evaluation cannot be treated as a privacy guarantee.

\section{From Behavioral Forgetting to Privacy-Oriented Evaluation}
\label{sec:privacy_evaluation}

\subsection{Current Unlearning Evaluation}

LLM unlearning benchmarks typically evaluate forgetting on a designated forget set while measuring preservation of unrelated knowledge or broader model capabilities. Representative benchmarks include TOFU \citep{maini2024tofutaskfictitiousunlearning}, WMDP \citep{li2024wmdpbenchmarkmeasuringreducing}, and RWKU \citep{jin2024rwkubenchmarkingrealworldknowledge}.

TOFU exemplifies this paradigm, evaluating whether an unlearned model resembles a retain-only reference model on synthetic factual associations via a Truth Ratio and KS-test criterion ($p>0.1$)~\citep{maini2024tofutaskfictitiousunlearning}. Model utility is measured using the harmonic mean of nine capability metrics, with results reported relative to the corresponding base model.

\subsection{The Evaluation Gap}

These metrics establish \emph{behavioral forgetting} under a predefined evaluation distribution, but provide limited evidence about whether the same information remains accessible elsewhere. This concern arises across languages \citep{lu2025learnunlearnaddressingmisinformation}, modalities \citep{patil2025unlearningsensitiveinformationmultimodal,chen2025safeeraserenhancingsafetymultimodal}, and adversarial interactions designed to recover residual information \citep{patil2025unlearningsensitiveinformationmultimodal,zhang2025suastealthymultimodallarge}. Thus, satisfying a benchmark forgetting criterion should not by itself be interpreted as evidence that targeted information is inaccessible.

For privacy-oriented evaluation, we therefore treat \emph{information inaccessibility} as a stronger objective and ask how much empirical evidence can be obtained that forgetting persists beyond the original evaluation distribution. Rather than claiming to establish inaccessibility, this shifts evaluation toward testing the robustness of forgetting across alternative mechanisms for accessing the same targeted information.

\subsection{Access-Path Evaluation}

We refer to these alternative mechanisms as \emph{access paths}. An access path is a transformation, interaction mechanism, or recovery procedure through which the same underlying targeted information may be accessed. Examples include paraphrases, translations, alternative scripts or modalities, encoded or obfuscated queries, and adversarial extraction procedures. Privacy-oriented evaluation should characterize whether conclusions about forgetting remain robust as these access paths change.

Cross-lingual evaluation provides a particularly controlled example. Given semantically equivalent queries in multiple languages, unlearning can be performed in one language and evaluated in another while holding the targeted factual information approximately fixed. This allows us to test whether apparent forgetting generalizes beyond the linguistic representation in which unlearning was performed.

\section{Cross-Lingual Unlearning as a Privacy Stress Test}
\label{sec:crosslingual}

\subsection{Experimental Setting}

We use \textbf{cTOFU}~\citep{lizzo2026evaluatingcrosslingualunlearningmultilingual}, a multilingual extension of TOFU \citep{maini2024tofutaskfictitiousunlearning}, to study the linguistic access path directly. cTOFU provides parallel versions of TOFU's synthetic factual associations in English, Chinese, Hindi, Italian, and Spanish, together with romanized Hindi and Chinese, while preserving the original forget/retain partitions. The use of synthetic facts does not itself model sensitive personal information, but provides a controlled setting in which the targeted information can be held fixed while its linguistic realization changes, isolating the effect of this one access path without requiring the use of real sensitive data.

We focus on \emph{one-to-one unlearning}: unlearning is performed in one language and the resulting model is evaluated in another. All experiments use the 10\% forget set and the TOFU forget-quality and normalized model-utility metrics described in Section~\ref{sec:privacy_evaluation}. Figure~\ref{fig:crosslingual_privacy} presents Llama~3 with \textsc{UNLEARN}~\citep{lizzo2025unlearn} as a representative case from the broader cTOFU study (Appendix~\ref{app:method-comparison}: baseline comparison; Appendix~\ref{app:model-coverage}: model coverage).

\subsection{Forgetting Depends on the Linguistic Access Path}

\begin{figure}[t]
    \centering
    \includegraphics[width=0.8\linewidth]
        {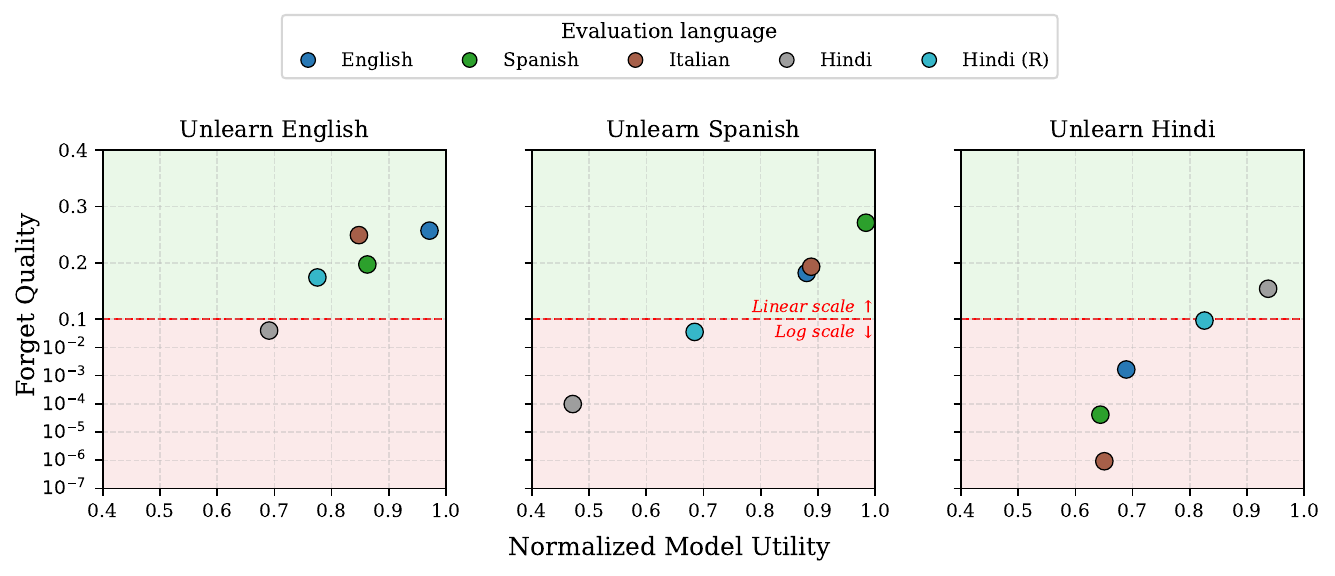}
    \caption{
        Cross-lingual access-path evaluation for Llama~3 using \textsc{UNLEARN}~\citep{lizzo2025unlearn} Each panel fixes the unlearning language, while points evaluate the same targeted facts through different linguistic access paths. The dashed line denotes the TOFU forgetting criterion ($p=0.1$); model utility is normalized relative to the base model. Apparent unlearning success varies with the language used for evaluation. See Appendix~\ref{app:additional-results} for additional results across model families.
    }
    \label{fig:crosslingual_privacy}
\end{figure}

Figure~\ref{fig:crosslingual_privacy} shows that conclusions about forgetting depend strongly on the linguistic access path used for evaluation. After unlearning in English, the model satisfies the TOFU forgetting criterion when evaluated in English, Spanish, Italian, and romanized Hindi, but fails when the same targeted information is evaluated in native-script Hindi. The reverse direction is even more pronounced: after unlearning in Hindi, forgetting succeeds in Hindi but does not transfer consistently to the other evaluation languages. These differences occur alongside substantial variation in retained model utility (additional results are provided in Appendix~\ref{app:additional-results}).

The broader cTOFU experiments exhibit similar dependence on language and script across multilingual models. In particular, cross-lingual forgetting is asymmetric, and native-script Hindi and Chinese generally exhibit weaker forgetting and utility than their romanized counterparts (Table~\ref{tab:additional-results}). Thus, even the narrowest kind of access-path variation, a meaning-preserving change in linguistic form, is already enough to change whether an unlearning operation appears successful.

\subsection{Representational Basis of Access-Path Dependence}
\label{sec:forgetting_language}

\begin{figure}[t]
    \centering
    \includegraphics[
        width=\linewidth,
        trim=0cm .15cm 0cm 0.15cm,
        clip
    ]{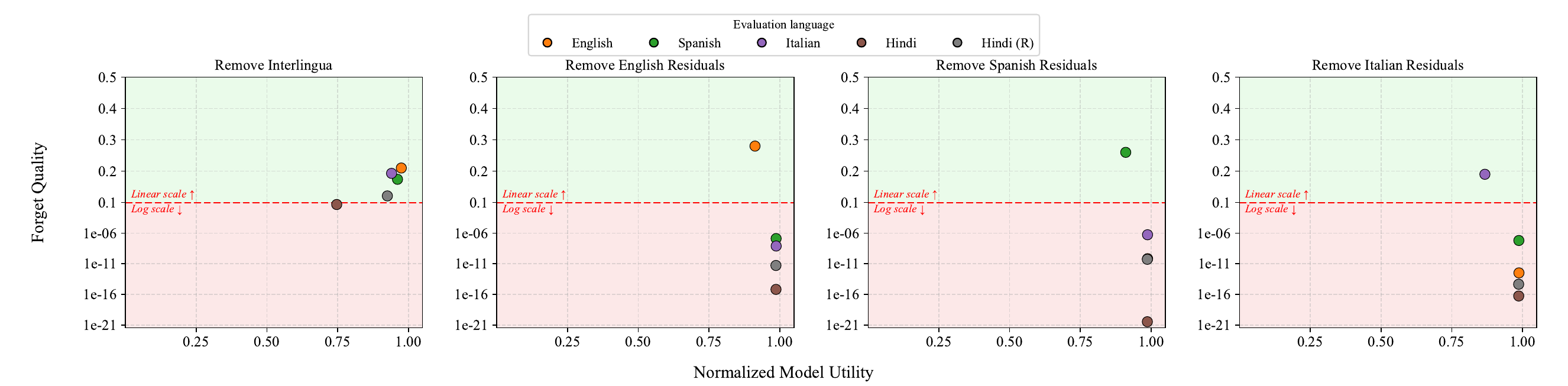}
    \caption{
        Representational basis of access-path-dependent forgetting. Removing the shared multilingual subspace induces forgetting across linguistic access paths, whereas removing a language-specific residual primarily affects the corresponding language.
    }
    \label{fig:interlingua_removal}
\end{figure}

The cTOFU representation analysis provides a mechanistic explanation, within this single access path, for why different linguistic access paths can produce different conclusions about forgetting. Task-specific weight subspaces exhibit substantial overlap across languages, consistent with factual knowledge being represented through a shared semantic component together with smaller language-specific residuals. Figure~\ref{fig:interlingua_removal} provides intervention evidence for this decomposition: removing the shared multilingual subspace induces forgetting across languages, whereas removing a language-specific residual primarily affects the targeted language.

These interventions indicate that cross-lingual differences are not solely artifacts of translated prompts: semantically equivalent facts can involve both shared and language-dependent representational structure (Appendix~\ref{app:translation-validation}: translation validation; Appendix~\ref{app:unlearn-method}: method details). Consequently, success under one linguistic access path need not imply comparable forgetting under another.

\section{Toward Privacy-Oriented Unlearning Evaluation}
\label{sec:privacy_framework}

Cross-lingual querying is a controlled stress test of one access path, and it is already enough to show that a model can satisfy a fixed forgetting criterion while targeted information remains accessible through another access path. Linguistic transformations are only one family of access mechanisms; the same information may also be accessed through different formulations, modalities, or interactions. If forgetting is this fragile under one meaning-preserving transformation, a privacy-oriented standard cannot stop at one benchmark distribution, and future work should extend testing across the broader space of access paths.

Future benchmarks should evaluate three increasingly demanding levels. \textbf{(1) In-distribution forgetting} tests the original forget examples and benchmark distribution; this is the current default. \textbf{(2) Access-path generalization} tests whether forgetting persists under meaning-preserving changes such as paraphrases, translations, alternative scripts, or modalities; our results already demonstrate failures here. \textbf{(3) Adversarial recoverability} asks whether targeted information can be recovered through deliberately constructed prompts, extraction, relearning, or other attacks; prior work shows such failures can emerge even when levels 1--2 are satisfied~\citep{patil2025unlearningsensitiveinformationmultimodal, zhang2025suastealthymultimodallarge}, making this a natural next target.

Future benchmarks should therefore distinguish \emph{forgetting on the target distribution} from \emph{robustness across alternative access paths}, with relevant paths depending on the privacy claim (paraphrases, languages, modalities, adversarial prompting, extraction attacks, or white-box recovery). Such evaluation cannot certify deletion, but can provide far stronger evidence that apparent forgetting reflects reduced accessibility rather than suppression under a particular benchmark distribution.

\section{Conclusion}
Behavioral forgetting on a designated benchmark distribution does not establish that targeted information is inaccessible: our cross-lingual case study shows conclusions about unlearning success can change within a single, controlled access path. If forgetting is this fragile under one meaning-preserving transformation, a fixed benchmark distribution is not a reliable privacy standard. We recommend that benchmarks report which access paths a forgetting claim has been tested against, extend testing beyond language, and treat adversarial recoverability as a priority for future benchmark design.

\section*{Limitations}

Our main cross-lingual analysis focuses on a single representative unlearning method (UNLEARN); while Appendix~\ref{app:method-comparison} shows this method outperforms several standard baselines in the monolingual setting, we do not claim that the specific magnitude of cross-lingual forgetting failure reported here generalizes to all unlearning algorithms. Second, our case study uses synthetic factual associations from TOFU and its multilingual extension cTOFU; these provide a controlled setting for isolating linguistic access-path effects, but do not establish that the same dynamics hold for real personal or sensitive information embedded during pretraining, which may be represented differently than fictional facts introduced during fine-tuning. Third, our evaluation spans five languages, including native-script and romanized representations of Hindi and Chinese; this is not intended as a claim of universal linguistic coverage, and forgetting behavior for typologically distant or lower-resource languages remains untested.

\section*{Acknowledgements}
This work was supported by CoCoSys, one of seven centers in JUMP 2.0, a Semiconductor Research Corporation (SRC) program sponsored by DARPA.

\bibliographystyle{plainnat}
\bibliography{neurips_2026}

\appendix

\section{Supplementary Materials}
\label{app:supplementary}

This appendix provides additional methodological detail and supporting results referenced in Sections~\ref{sec:privacy_evaluation}--\ref{sec:privacy_framework} of the main text. These materials are drawn from the broader cross-lingual unlearning study (cTOFU) from which the results in Figure~\ref{fig:crosslingual_privacy} and Figure~\ref{fig:interlingua_removal} are drawn~\citep{lizzo2026evaluatingcrosslingualunlearningmultilingual}, and from the original UNLEARN method paper~\citep{lizzo2025unlearn}.

\subsection{Why UNLEARN: Comparison Against Baseline Unlearning Methods}
\label{app:method-comparison}

Figure~\ref{fig:crosslingual_privacy} and Figure~\ref{fig:interlingua_removal} in the main text report results for UNLEARN~\citep{lizzo2025unlearn} as a representative unlearning method. This choice is motivated by its consistently stronger forget-quality/utility trade-off relative to standard baselines (DPO, Gradient Ascent, KL Minimization, Gradient Difference, NPO, and FLAT) in the monolingual (English-English) setting, prior to introducing any cross-lingual access-path variation. Table~\ref{tab:method-comparison} reports this monolingual baseline comparison across three multilingual model families.

\begin{table*}[h]
\centering
\small
\begin{tabular}{lcccccc}
\toprule
& \multicolumn{2}{c}{\textbf{Llama 3}} 
& \multicolumn{2}{c}{\textbf{Mixtral}} 
& \multicolumn{2}{c}{\textbf{Aya 23}} \\
\cmidrule(lr){2-3}\cmidrule(lr){4-5}\cmidrule(lr){6-7}\\
\textbf{Method} 
& Utility $\uparrow$ & Forget $\uparrow$ 
& Utility $\uparrow$ & Forget $\uparrow$ 
& Utility $\uparrow$ & Forget $\uparrow$ \\
\midrule
Base Model                & 0.6227 & $2.37{\times}10^{-15}$ & 0.5912 & $3.0{\times}10^{-16}$ & 0.5503 & $6.4{\times}10^{-24}$ \\
\midrule
DPO  & 0.5357 & $1.06{\times}10^{-16}$ & 0.5100 & $2.5{\times}10^{-14}$ & 0.4182 & $4.3{\times}10^{-21}$ \\
Gradient Ascent           & 0.0000 & $1.43{\times}10^{-22}$ & 0.0000 & $1.35{\times}10^{-19}$ & 0.0000 & $6.28{\times}10^{-18}$ \\
KL Minimization           & 0.0000 & $4.07{\times}10^{-32}$ & 0.0000 & $7.23{\times}10^{-19}$ & 0.0000 & $2.38{\times}10^{-26}$ \\
Gradient Difference       & 0.5872 & $2.83{\times}10^{-25}$ & 0.5483 & $5.83{\times}10^{-25}$ & 0.4978 & $6.19{\times}10^{-29}$ \\
NPO                       & 0.2297 & 0.0291                & 0.1893 & 0.0587                & 0.1213 & 0.0468 \\
FLAT                      & \textbf{0.6104} & 0.0841                & \textbf{0.5871} & 0.0692                & \textbf{0.5264} & 0.0765 \\
UNLEARN                   & \underline{0.6047} & \textbf{0.257} \checkmark        & \underline{0.5741} & \textbf{0.215} \checkmark        & \underline{0.5047} & \textbf{0.315} \checkmark \\
\bottomrule
\end{tabular}
\caption{Comparison of unlearning performance on the English-English TOFU benchmark \citep{maini2024tofutaskfictitiousunlearning} for three models: Llama 3, Mixtral, and Aya 23. Among unlearning methods, the best value in ecah column is bolded and the second-best is underlined. \checkmark marks forget quality above $\alpha=0.1$.}
\label{tab:method-comparison}
\end{table*}
Across all three model families, UNLEARN is the only method that satisfies the TOFU forgetting criterion ($p>0.1$) while maintaining normalized utility close to that of the base model. This motivates its use as the representative method for the cross-lingual access-path evaluation presented in the main text, rather than a method selected post hoc to support a particular narrative.

\subsection{Translation Validation for cTOFU}
\label{app:translation-validation}

The cross-lingual access paths evaluated in Section~\ref{sec:crosslingual} rely on the cTOFU dataset~\citep{lizzo2026evaluatingcrosslingualunlearningmultilingual}, a multilingual extension of TOFU~\citep{maini2024tofutaskfictitiousunlearning} spanning Spanish, Italian, Hindi (native script and romanized), and Chinese (native script and pinyin). To confirm that forgetting differences across languages reflect genuine cross-lingual access-path variation rather than translation artifacts, cTOFU's translations were validated using both automatic and human evaluation. Table~\ref{tab:translation-validation} reports COMET-Kiwi reference-free quality estimation scores and human pass rates (400 randomly sampled examples per language, evaluated by fluent bilingual reviewers for preservation of author identity, factual associations, answer correctness, and semantic consistency).

\begin{table}[H]
\centering
\small
\begin{tabular}{l c c}
\toprule
\textbf{Language} & \textbf{Human Pass Rate} & \textbf{COMET-Kiwi}  \\
\midrule
Spanish & 98.75\% & 0.8818 \\
Hindi & 96.5\%& 0.8769 \\
Hindi (Romanized) & 94.25\%& -- \\
Italian & 99.25\% & 0.8858 \\
Chinese & 95.5\% & 0.8610 \\
Chinese (Pinyin) & 93.25\% & -- \\
\bottomrule
\end{tabular}
\caption{Translation validation results for cTOFU. COMET-Kiwi scores measure semantic consistency between English source text and translated outputs using a reference-free quality estimation model. Human validation pass rates report the percentage of manually inspected examples preserving factual consistency and author identity across translations. Human evaluation used 400 randomly sampled examples per language.}
\label{tab:translation-validation}
\end{table}

All automatically scored languages exceed a COMET-Kiwi score of 0.85, and all languages exceed a 93\% human validation pass rate, indicating that the semantic content of the targeted factual associations is preserved across translation and transliteration. This supports the interpretation in Section~\ref{sec:forgetting_language} that cross-lingual differences in forgetting reflect genuine representational structure rather than translation noise.

\subsection{UNLEARN: Method Details}
\label{app:unlearn-method}

The representative unlearning method used throughout the cross-lingual evaluation, UNLEARN~\citep{lizzo2025unlearn}, consists of three stages: subspace identification, subspace discrimination, and task removal.

\paragraph{Subspace identification.} UNLEARN identifies the subspace of a targeted task (here, a language-specific or shared factual association) within the weight space of the LLM via sequential, layer-by-layer training. Starting from randomly initialized weights, each layer $l$ is trained in isolation while all other layers remain frozen, maximizing the conditional language modeling objective on the targeted task data. To accommodate the rank-deficient nature of the resulting task matrix, each layer is parameterized with a low-rank bottleneck architecture $T_i^l = FG$ with interior dimension $k$. This layer-by-layer procedure yields a set of matrices $T_i^l$ representing the task-dependent subspace at each layer, with early layers capturing higher-level semantic structure and later layers capturing more task-specific detail.

\paragraph{Subspace discrimination.} Naively subtracting the identified task subspace can degrade performance on related tasks that occupy overlapping regions of weight space. UNLEARN addresses this using a variation of the Gram–Schmidt orthogonalization process, which removes only the information unique to the targeted task while preserving shared (mutual) information with related tasks. For a target task subspace $T_i$ and one or more related task subspaces $T_o$, the discrimination step computes, for each singular vector,
\[
SV_k(T_i') = SV_k(T_i) - \sum_{j=1}^{N} \frac{SV_k(T_i)\cdot SV_j(T_o)}{SV_j(T_o)\cdot SV_j(T_o)}\, SV_j(T_o),
\]
yielding a modified task matrix $T_i'$ that is orthogonal to the related-task subspace $T_o$. This step is central to the cross-lingual analysis in Section~\ref{sec:forgetting_language}: applying this discrimination procedure separately to the shared multilingual subspace and to each language-specific residual subspace enables the intervention ablations reported in Figure~\ref{fig:interlingua_removal}.

\paragraph{Task removal.} The final step reconstitutes the modified task matrix $T_i'$ via SVD and subtracts a (possibly smoothed) combination of $T_i'$ and the original task matrix $T_i$ from the model weights:
\[
W' = W - \alpha T_i' - (1-\alpha) T_i,
\]
where $\alpha \in [0,1]$ trades off forgetting strength against preservation of related-task performance. In the interlingua intervention experiments (Section~\ref{sec:forgetting_language}), this mechanism is applied twice: once with the shared multilingual subspace as the targeted component, and once with each language-specific residual subspace as the targeted component, isolating the causal contribution of each to cross-lingual forgetting behavior.

\subsection{Model Coverage}
\label{app:model-coverage}

Table~\ref{tab:model-coverage} lists the pretrained model checkpoints evaluated across the cTOFU experiments summarized in the main text, spanning multiple generations and degrees of multilingual pretraining coverage.

\begin{table}[H]
\centering
\small
\begin{tabular}{l l l l}
\toprule
\textbf{Model} & \textbf{Active Parameters} & \textbf{Total Parameters} & \textbf{HuggingFace Checkpoint} \\
\midrule
Mistral 7B & 7B & 7B & \texttt{mistralai/Mistral-7B-v0.1} \\
Mixtral 8$\times$7B & 12.9B & 46.7B & \texttt{mistralai/Mixtral-8x7B-v0.1} \\
Magistral Small & 24B & 24B & \texttt{mistralai/Magistral-Small-2506} \\
LLaMA 2 7B & 7B & 7B & \texttt{meta-llama/Llama-2-7b-hf} \\
LLaMA 3 8B & 8B & 8B & \texttt{meta-llama/Meta-Llama-3-8B} \\
LLaMA 4 Scout & 17B & 109B & \texttt{meta-llama/Llama-4-Scout-17B-16E} \\
Aya 23 8B & 8B & 8B & \texttt{CohereForAI/aya-23-8B} \\
\bottomrule
\end{tabular}
\caption{Pretrained language models used in our evaluation. Active parameters denote the number of parameters used during a forward pass. Total parameters denote the full parameter count of the model. All checkpoints are obtained from HuggingFace.}
\label{tab:model-coverage}
\end{table}

Not all models were evaluated on every language; language coverage was restricted to languages represented during each model's multilingual pretraining (e.g., Llama-family models were not evaluated on Chinese, and Mixtral was not evaluated on Hindi), so that differences in cross-lingual forgetting reflect the underlying unlearning procedure rather than confounds from unsupported language capability.

\newpage
\subsection{Additional Cross-Lingual Results}
Table~\ref{tab:additional-results} reports the full one-to-one cross-lingual unlearning results for Llama~3, Mixtral, and Aya~23. The results reinforce the access-path dependence observed in the main text. Forgetting often transfers among closely related linguistic representations, but can degrade substantially across languages and scripts, and the direction of transfer is asymmetric. These results demonstrate that the behavior shown for Llama~3 in Figure~\ref{fig:crosslingual_privacy} is not unique to a single model family, although the specific pattern and magnitude of cross-lingual transfer vary across models.

\label{app:additional-results}

\begin{table}[H]
\centering
\scriptsize
\setlength{\tabcolsep}{4pt}
\begin{tabular}{ll rr rr rr}
\toprule
& & \multicolumn{2}{c}{\textbf{Llama 3}}
  & \multicolumn{2}{c}{\textbf{Mixtral}}
  & \multicolumn{2}{c}{\textbf{Aya 23}} \\
\cmidrule(lr){3-4}\cmidrule(lr){5-6}\cmidrule(lr){7-8}
Unlearn & Eval & Util. & FQ & Util. & FQ & Util. & FQ \\
\midrule

\multirow{7}{*}{Base}
& English     & 0.6227 & $2.37{\times}10^{-15}$ & 0.5912 & $2.74{\times}10^{-16}$ & 0.5503 & $6.47{\times}10^{-24}$ \\
& Spanish     & 0.5919 & $4.31{\times}10^{-13}$ & 0.5393 & $4.68{\times}10^{-13}$ & 0.5193 & $2.63{\times}10^{-21}$ \\
& Italian     & 0.6102 & $8.84{\times}10^{-17}$ & 0.5669 & $9.31{\times}10^{-19}$ & 0.5117 & $3.21{\times}10^{-19}$ \\
& Hindi       & 0.4104 & $7.67{\times}10^{-17}$ & --     & --                     & 0.3929 & $5.73{\times}10^{-21}$ \\
& Hindi (R)   & 0.5293 & $7.88{\times}10^{-12}$ & --     & --                     & 0.4623 & $3.51{\times}10^{-18}$ \\
& Chinese     & --     & --                     & 0.0507 & $3.79{\times}10^{-12}$ & 0.2688 & $4.62{\times}10^{-19}$ \\
& Chinese (R) & --     & --                     & 0.1693 & $1.95{\times}10^{-18}$ & 0.2983 & $2.82{\times}10^{-23}$ \\

\midrule
\multirow{7}{*}{English}
& English     & 0.6047 & 0.257\,\checkmark & 0.5741 & 0.215\,\checkmark & 0.5047 & 0.315\,\checkmark \\
& Spanish     & 0.5105 & 0.197\,\checkmark & 0.4560 & 0.177\,\checkmark & 0.4210 & 0.248\,\checkmark \\
& Italian     & 0.5173 & 0.249\,\checkmark & 0.4527 & 0.201\,\checkmark & 0.4513 & 0.263\,\checkmark \\
& Hindi       & 0.2835 & 0.0396            & --     & --                & 0.2550 & 0.0398 \\
& Hindi (R)   & 0.4103 & 0.174\,\checkmark & --     & --                & 0.3610 & 0.149\,\checkmark \\
& Chinese     & --     & --                & 0.00647 & $1.38{\times}10^{-11}$ & 0.1460 & 0.0457 \\
& Chinese (R) & --     & --                & 0.0503 & $5.38{\times}10^{-4}$  & 0.2420 & 0.153\,\checkmark \\

\midrule
\multirow{7}{*}{Spanish}
& English     & 0.5482 & 0.182\,\checkmark & 0.4920 & 0.162\,\checkmark & 0.4073 & 0.192\,\checkmark \\
& Spanish     & 0.5824 & 0.271\,\checkmark & 0.4942 & 0.241\,\checkmark & 0.4621 & 0.272\,\checkmark \\
& Italian     & 0.5420 & 0.193\,\checkmark & 0.4757 & 0.0394            & 0.4230 & 0.178\,\checkmark \\
& Hindi       & 0.1935 & $9.83{\times}10^{-5}$ & -- & --               & 0.1647 & 0.00129 \\
& Hindi (R)   & 0.3623 & 0.0354            & --     & --                & 0.3954 & 0.0647 \\
& Chinese     & --     & --                & 0.00307 & $9.83{\times}10^{-19}$ & 0.0829 & $6.46{\times}10^{-5}$ \\
& Chinese (R) & --     & --                & 0.00124 & $1.37{\times}10^{-10}$ & 0.1865 & 0.0826 \\

\midrule
\multirow{7}{*}{Italian}
& English     & 0.5930 & 0.216\,\checkmark & 0.5461 & 0.152\,\checkmark & 0.5160 & 0.229\,\checkmark \\
& Spanish     & 0.5203 & 0.195\,\checkmark & 0.4785 & 0.165\,\checkmark & 0.4534 & 0.183\,\checkmark \\
& Italian     & 0.5623 & 0.284\,\checkmark & 0.4967 & 0.238\,\checkmark & 0.4912 & 0.291\,\checkmark \\
& Hindi       & 0.2164 & $1.63{\times}10^{-4}$ & -- & --               & 0.1373 & 0.00367 \\
& Hindi (R)   & 0.3319 & 0.0137            & --     & --                & 0.3521 & 0.0517 \\
& Chinese     & --     & --                & 0.00219 & $1.27{\times}10^{-22}$ & 0.1250 & $8.32{\times}10^{-4}$ \\
& Chinese (R) & --     & --                & 0.00216 & $5.67{\times}10^{-12}$ & 0.1565 & 0.0659 \\

\midrule
\multirow{7}{*}{Hindi}
& English     & 0.4290 & 0.00165           & --     & --                & 0.3810 & 0.00837 \\
& Spanish     & 0.3810 & $4.13{\times}10^{-5}$ & -- & --               & 0.3130 & $6.12{\times}10^{-6}$ \\
& Italian     & 0.3970 & $9.23{\times}10^{-7}$ & -- & --               & 0.2980 & $1.73{\times}10^{-7}$ \\
& Hindi       & 0.3845 & 0.154\,\checkmark & --     & --                & 0.3194 & 0.194\,\checkmark \\
& Hindi (R)   & 0.4370 & 0.0896            & --     & --                & 0.3852 & 0.156\,\checkmark \\

\midrule
\multirow{7}{*}{Hindi (R)}
& English     & 0.5104 & 0.112\,\checkmark & --     & --                & 0.4521 & 0.134\,\checkmark \\
& Spanish     & 0.4670 & 0.0716            & --     & --                & 0.3950 & 0.0638 \\
& Italian     & 0.4850 & 0.0517            & --     & --                & 0.3540 & 0.0341 \\
& Hindi       & 0.3728 & 0.153\,\checkmark & --     & --                & 0.3152 & 0.178\,\checkmark \\
& Hindi (R)   & 0.4823 & 0.187\,\checkmark & --     & --                & 0.4211 & 0.210\,\checkmark \\

\midrule
\multirow{7}{*}{Chinese}
& English     & --     & --                & 0.0114 & $4.65{\times}10^{-9}$  & 0.3120 & $5.26{\times}10^{-4}$ \\
& Spanish     & --     & --                & 0.00446 & $1.30{\times}10^{-11}$ & 0.2560 & $9.85{\times}10^{-9}$ \\
& Italian     & --     & --                & 0.0110 & $1.92{\times}10^{-13}$ & 0.2150 & $9.33{\times}10^{-14}$ \\
& Chinese     & --     & --                & 0.0178 & $1.24{\times}10^{-5}$  & 0.2188 & 0.124\,\checkmark \\
& Chinese (R) & --     & --                & 0.0813 & $5.93{\times}10^{-5}$  & 0.2250 & 0.0854 \\

\midrule
\multirow{7}{*}{Chinese (R)}
& English     & --     & --                & 0.3731 & 0.00734           & 0.3160 & 0.0863 \\
& Spanish     & --     & --                & 0.2679 & 0.00374           & 0.2680 & 0.0643 \\
& Italian     & --     & --                & 0.2416 & 0.00131           & 0.2320 & 0.0394 \\
& Chinese     & --     & --                & 0.0128 & $1.37{\times}10^{-6}$  & 0.1865 & 0.116\,\checkmark \\
& Chinese (R) & --     & --                & 0.1284 & 0.0396            & 0.2578 & 0.184\,\checkmark \\

\bottomrule
\end{tabular}
\caption{Full one-to-one cross-lingual unlearning results across \textbf{Llama 3}, \textbf{Mixtral}, and \textbf{Aya 23}. Rows indicate the evaluation language within each unlearning-language block. A $\checkmark$ indicates satisfaction of the TOFU forgetting criterion ($p>0.1$).}
\label{tab:additional-results}
\end{table}

\end{document}